\documentclass[runningheads]{llncs}
\usepackage{graphicx}
\usepackage[linesnumbered,ruled,vlined]{algorithm2e}
\SetKwInOut{Required}{Required}

\usepackage{amsfonts}
\usepackage{amsmath} 
\usepackage[english]{babel}
\usepackage{tabularx} 
\usepackage{booktabs} 
\usepackage{multirow}
\usepackage{subcaption}
\usepackage{xcolor}
\begin{document}

\title{A Prompt-driven Universal Model for View-Agnostic Echocardiography Analysis}
% sekeun, kyungsang, Guo peng, Abder rahman, eugenee, chen cheng, xiang li 
%\titlerunning{Abbreviated paper title}
% If the paper title is too long for the running head, you can set
% an abbreviated paper title here
%
\author{Sekeun Kim\inst{1} \and
Hui Ren\inst{1} \and
Peng Guo\inst{1} \and
Abder-Rahman Ali\inst{1} \and
Patrick Zhang\inst{1} \and
Kyungsang Kim\inst{1} \and
Quanzheng Li\inst{1} \and
Xiang Li\inst{1}}

\institute{Massachusetts General Hospital and Harvard Medical School}

\maketitle        
\begin{abstract}
Echocardiography segmentation for cardiac analysis is time-consuming and resource-intensive due to the variability in image quality and the necessity to process scans from various standard views. While current automated segmentation methods in echocardiography show promising performance, they are trained on specific scan views to analyze corresponding data. However, this solution has a limitation as the number of required models increases with the number of standard views. To address this, in this paper, we present a prompt-driven universal method for view-agnostic echocardiography analysis. Considering the domain shift between standard views, we first introduce a method called prompt matching, aimed at learning prompts specific to different views by matching prompts and querying input embeddings using a pre-trained vision model. Then, we utilized a pre-trained medical language model to align textual information with pixel data for accurate segmentation. Extensive experiments on three standard views showed that our approach signiﬁcantly outperforms the state-of-the-art universal methods and achieves comparable or even better performances over the segmentation model trained and tested on same views. 

\end{abstract}

\section{Introduction}
Echocardiography is the most frequently used imaging modality performed in cardiology, which helps assessment of cardiac function by examining different views by multiple standard views scanning. Given the complexity in image analysis and workload of sonographers, there is a growing interest in developing automated methods for segmentation in Echo \cite{kim2022fully,kim2021automatic,leclerc2020lu}. Existing methods have demonstrated superior performance  in accurately delineating anatomical structures within specific views when they are trained on corresponding datasets. This process involves a step of identifying the desired views in the patient study before conducting the analysis, which requires an additional steps to select the appropriate files in the scan files \cite{charton2023multi,jeon2023improving}. The exploration of a general model capable of independently performing echocardiography segmentation tasks across multiple standard views has not been explored.\\
Currently, the general solution is training \textit{N} models on \textit{N} standard views. However, this solutions have a limitation as the number of models are increases with the number of standard views. One naive approach to creating a universal model is to train the same network on data from various standard view scans. This approach may result in a performance degradation of the model, as each standard view manifests its distinct visual characteristics \cite{kim2021automatic,mitchell2019guidelines}. Echocardiography poses distinct challenges, including view domain shift among scan views and sparse annotation across frames. While there may not be identical solutions to these problems, similar universal models have been developed \cite{zhang2021dodnet,butoi2023universeg,liu2023clip,ye2023uniseg}. DoDNet\cite{zhang2021dodnet} introduces a dynamic head with an encoder-decoder architecture. Different task information is encoded to one-hot vector, and it controls the task-specific controller. CLIP-drvien universal model\cite{liu2023clip} extends this idea by utilizing a pre-trained text model and managing segmentation heads using semantically embedded class features. Although this CLIP-based strategy has demonstrated success in CT organ segmentation tasks, it has limitations in the medical framework due to disparities between natural and medical texts. UniSeg\cite{ye2023uniseg} approaches with learnable prompts have been developed to address various segmentation tasks in CT,MR, and PET. However, it introduces a universal method using three anatomically similar datasets, where the images may differ only in texture while the underlying anatomy remains the same. Consequently, approach is likely to be less suitable for addressing view shifts in echocardiography, potentially resulting in suboptimal performance as in Table \ref{table2}.\\
To address above problems, we propose a prompt-driven universal model that allows superior segmentation of cardiac structures with state-of-the-art performance. Our model incorporates a prompt learning with prompt pool and pre-trained language model's knowledge by pixel-text alignment. We first employ a prompt pool-based prompt learning approach enables the development of a universal model capable of handling various scan view data. Accordingly, it effectively tackles the aforementioned issue by enabling dynamic adaptation to diverse inputs. Second, The score maps facilitates pixel-text alignment, which allows our model to fully leverage language information for medical segmentation tasks. It effectively address the above problem by enabling input video to select particular view specific prompt and focus on particular semantic features with language model. To the best of our knowledge, this is the first work that approaches unified model segmentation in echocardiography. Our method simplifies current cardiac analysis by removing the necessity for a view identification step to acquire the desired view from various DICOM scans of the patient. Our method is evaluated on three standard views from three different datasets and show promising performance compared to other universal methods. 
\\
Our contributions can be summarized as follows: \\
• We present a prompt-driven universal model, comprising a prompt pool to accommodate different standard views, and leveraging pixel-text alignment with the prior knowledge of a pre-trained text model for view-agnostic echocardiography segmentation.\\
• The proposed method streamlines cardiac analysis by minimizing the requirement for a view identification step during the retrieval of the desired view from patient scans.\\
• We demonstrate that our model achieves SOTA performance for cardiac segmentation tasks compared to the previous universal approach through extensive experiments on various datasets. 

\begin{figure}[t]
    \centering
    \includegraphics[width=1.0\linewidth]{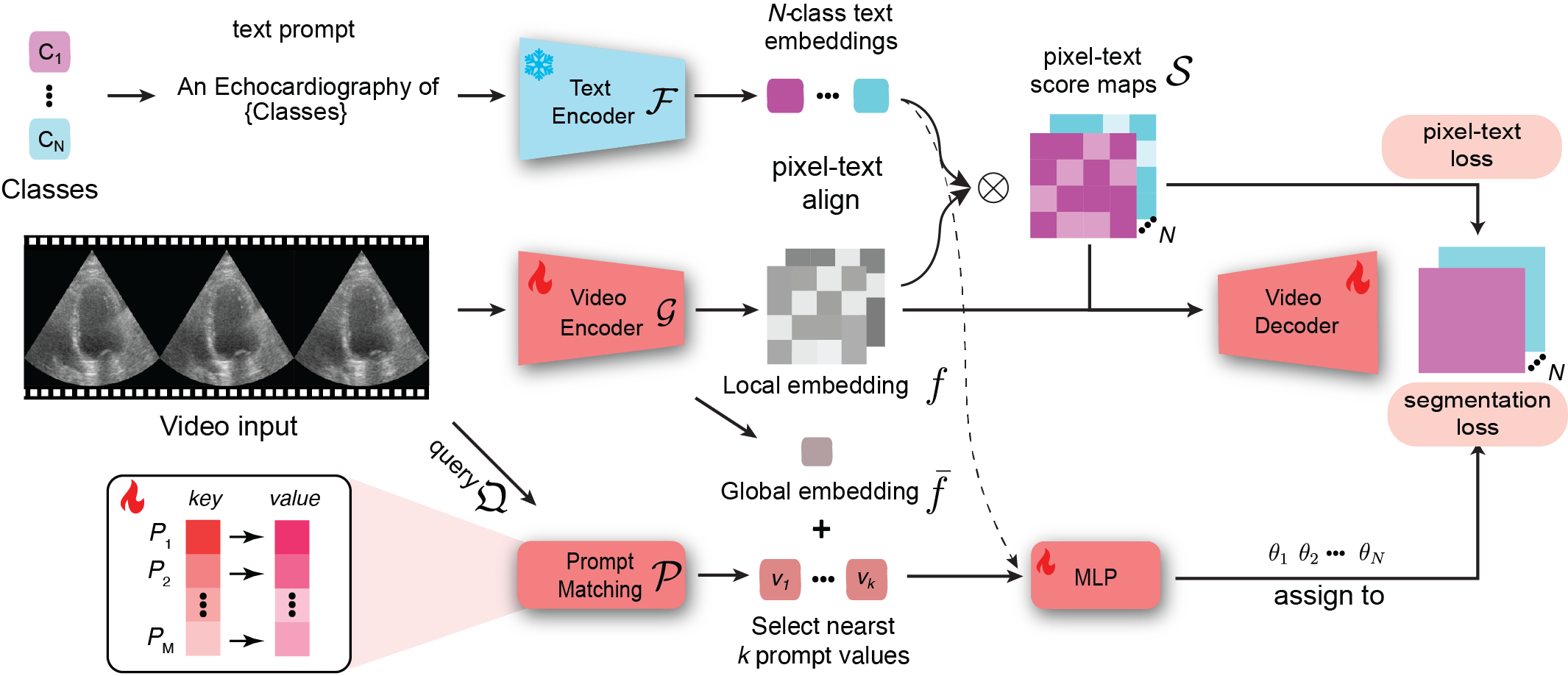}
    \caption{Over all framework of our proposed universal model for view-agnostic segmentation. The pre-trained model and query model remain frozen, while the other models are trainable. }
    \label{method}
\end{figure}

\section{Method}
As illustrated in Figure \ref{method}, our proposed approach comprises the following components: text, video encoder, a prompt pool consisting of trainable key and value, an MLP layer, and a video decoder. We utilize a clinicalBert \cite{alsentzer2019publicly} for enhanced extraction of medical text representations. Our goal is to segment objects in all frames across different scan views. In order to achieve this goal, we introduce two key components in our model: 1) a pixel-text dense alignment mechanism that effectively bridges the gap between a pre-trained language model and pixel-level representations for dense prediction tasks, and 2) a prompt matching technique that leverages a prompt pool to adaptively select the optimal view-specific prompt for each task.

\subsection{Problem Definition}
Given total number of \textit{N} datasets \textit{D} = $\{D_1, D_2, \ldots, D_N\}$, each dataset $D_{i} = \left \{ X_{ij}, Y_{ij} \right \}_{j=1}^{n_{i}}$, $X_{ij}$ and $Y_{ij}$ are the video with total frame of ${F}$ and the corresponding ground truth, represents total of $n_{i}$ pixels. Each videos $X_{ij} \in V_{k}$ , \textit{V} with total number of \textit{K} views $V \in \{V_1, V_2, \ldots, V_K\}$. If there are $\forall \textit{F}$ annotated in $Y_{ij}$,  $D_{i}$ is fully labeled dataset; otherwise $D_{i}$ is a partially labeled dataset. The objective is to train a model F(·) using the partially labeled dataset $D_{i}$ = $\{D_1, D_2, \ldots, D_N\}$, where the model is capable of making dense predictions for all K classes along all frame ${F}$.

\subsection{Pixel-text dense alignment}
In the computer vision field, a series of works on vision and language models (VLM) have emerged. Further in medical domain, previous studies have successfully adapted CLIP embeddings for medical applications \cite{qin2022medical,liu2023clip}. However, the use of CLIP trained on natural image-text pairs weakens the semantic meaning embedding for medical prompt into the models as shown in Table \ref{table3}. To fully leverage the knowledge encoded in the pre-trained medical language model, we utilized ClinicalBert \cite{alsentzer2019publicly} for dense prediction. We generate text embeddings by converting $N$ classes into text prompts using a template of ``An echocardiography of [Classes].'' to generate text embeddings $\mathcal{F}(c) \in \mathbb{R}^{N \times  D}$. An input video is encoded through the backbone video encoder to embed intermediate local video embedding $\mathcal{G}(x) \in \mathbb{R}^{T_{i}H_{i}W_{i} \times D} $, i = 1, ..., L where $T_{i}$, $H_{i}$, and $W_{i}$ are the frame, height and width of the local embeddings from i-th layer and D indicates embedding dimension. Then we compute the score maps with pixel-text alignment using the text embedding and the vision embedding by:
\[\mathcal{S} = \bar{\mathcal{G}(x)} \bar{\mathcal{F}(c)^{T}} \]
where superscript $^{-}$ refers the normalization along channel dimension and $^{T}$ denotes the transpose operation. The score map $\mathcal{S}$ can be employed for auxiliary segmentation at a lower resolution denoted as pixel-text loss. We concatenate pixel-text score map $\mathcal{S}$ to local embeddings ${f}$ to incorporate text priors. We utilized the chamber class for the text encoder without incorporating any view information.

\subsection{Prompt matching and text-driven parameter generation} 
 Given an 2D-t input $x \in \mathbb{R}^{T \times  H \times W \times C}$ and $\mathfrak{Q}$ is a pre-trained Vision Transformer(ViT) of Segment Anything Model \cite{kirillov2023segment}, first frame of video is divided into patches and then embed as patch embeddigs $\mathfrak{Q} : \mathbb{R}^{L \times (S^{2} \times C)}\rightarrow \mathbb{R}^{L \times D}$ where S indicates the patch size and C denotes input channels, and D is embedding dimension. The prompt pool consist of two learnable key $ \left \{ (k_1, P_1), (k_2, P_2), ... , (k_M, P_M) \right \}$, where $k_i \in \mathbb{R ^{D}}$ and learnable value $ \left \{ P_1, P_2, ..., P_M \right \}$, where $P_i \in \mathbb{R ^{L \times D}}$. In our settings, the total number of entries in the prompt pool, denoted by $M$, is equal to the number of views multiplied by a pre-defined prompt size assigned to each view, which is set to 3. The queried input embeddings and prompt keys for each views are drawn towards each other to maximize their cosine distance in training steps, denoted as \(\mathcal{L}_{pr}\). We use a global average pooling (GAP) layer on the last encoder features to obtain a global representation of the current video input. Then we utilize the text embeddings together with the prompt values and global embedding to generate parameters for the chamber segmentation heads, \(\theta_{N}\). These parameters used in video decoder heads and generate binary prediction for \({N}\) classess \cite{tian2020conditional}. This design facilitates view-agnostic prompt and preserving a view information during test time. 

\begin{table}[t]
    \centering
    \caption{Details of the publicly available datasets for training and evaluation.}
    \label{tab1}
    \begin{tabularx}{\textwidth}{
        >{\hsize=1.2\hsize}X % Dataset
        >{\centering\arraybackslash\hsize=0.7\hsize}X % View
        >{\centering\hsize=0.8\hsize}X % Annotation
        >{\centering\arraybackslash\hsize=1.0\hsize}X % #Total Scan (train/test)
    }
        \toprule
        Dataset & Scan View & Annotation & \#Total Scan (train/test) \\
        \midrule
        CAMUS \cite{leclerc2019deep} & A2C & LV$_{endo}$, LV$_{epi}$ & 500 (450/50)\\
         & A4C & LV$_{endo}$, LV$_{epi}$ & 500 (450/50) \\
        EchoNet-Pediatric \cite{reddy2023video} & A4C & LV$_{endo}$ & 3284 (2580/704) \\
        & PSAX & LV$_{endo}$ & 4526 (3559/967)  \\
        EchoNet-Dynamic \cite{ouyang2020video} & A4C & LV$_{endo}$ & 10036 (8753/1277) \\
        \bottomrule
    \end{tabularx}
\end{table}

\subsection{Loss Function}
\subsubsection{Video Masked Back-propagation}
In our problem,  the labels are distributed extremely sparsely across the frames which is difference from previous work \cite{liu2023clip}. In this work, we designed the video masked back propagation to address the partially labeled issue. Specifically, we masked frames that does not have class label and only back propagate loss to update parameter in our network. In this way, we can utilize sparsely labeled problem and do accurate segmentation in video with partially labeled dataset. 

\subsubsection{Total loss}
Our objective is to achieve segmentation by minimizing two loss terms including a prompt matching loss and segmentation loss with masked backpropagation through optimization of the following loss function:
\[ \mathcal{L}_{\text{seg}} = \lambda_{1}\mathcal{L}_{\text{pixel-text}} + \lambda_{2}\mathcal{L}_{\text{BCE}} , \quad \mathcal{L}_{\text{pr}} = < \mathfrak {Q} (X_{i0}), P_{key} >  \]
\[ \mathcal{L}_{\text{total}} =  (1 - \lambda(t)) \mathcal{L}_{\text{seg}} -  \lambda(t) \mathcal{L}_{\text{pr}} \]
where \(\mathcal{L}_{seg}\) denotes the segmentation loss combining different two losses, \(\mathcal{L}_{pixel-text}\) represents CE loss with score maps, and \(\mathcal{L}_{BCE}\) represents the binary cross-entropy loss, respectively. Throughout the experiments, \(\lambda_{1}\) and \(\lambda_{2}\) are set equally. \(\mathcal{L}_{pr}\) denotes the cosine similarity between queried input and prompt keys assigned by view types during training steps. \(\lambda\) is scheduled by the time dependent ramp up Gaussian function \(\lambda(t) = \exp^ {-5(1 - t / t_{max})^2}\) where t is the current iteration and $t_{max}$ is the maximal iteration. Since our prompt keys are converged in the early stage, we reduce the weights on prompt matching loss during this initial phase.

\begin{table}[t]
    \centering
    \caption{Quantitative comparison across all datasets. The Dice scores [\%] are presented, with the best results highlighted in \textbf{bold}.}
    \label{table2}
    \begin{tabularx}{\textwidth}{>{\raggedright}p{2.7cm} *{2}{>{\centering}X} *{2}{>{\centering}X} c *{3}{>{\centering\arraybackslash}X}}
        \toprule
        Method & \multicolumn{2}{c}{A2C} & \multicolumn{2}{c}{A4C} & \multicolumn{1}{c}{PSAX} & \multicolumn{3}{c}{Indiv A4C perf.} \\
        \cmidrule(lr){2-3} \cmidrule(lr){4-5} \cmidrule(lr){6-6} \cmidrule(lr){7-9}
         & LV$_{\text{endo}}$ & LV$_{\text{epi}}$ & LV$_{\text{endo}}$ & LV$_{\text{epi}}$ & LV$_{\text{endo}}$ & Camus & Pedia & Dynamic \\
        \midrule
        \textit{View-specific} & & & & & & & & \\
        SwinUNETR \cite{hatamizadeh2021swin} & 90.5 & 86.7 & 85.9 & 86.7 & 87.7 & 90.4 & 83.1 & 84.2 \\
        U-Transformer \cite{petit2021u} & \textbf{93.3} & 88.1 & 88.3 & \textbf{88.5} & 88.1 & 92.8 & \textbf{86.4} & 85.9\\
        
        \midrule
        \textit{View-integrated} & & & & & & & & \\
        SwinUNETR \cite{hatamizadeh2021swin} & 88.8 & 83.1 & 82.9 & 84.7 & 85.1 & 87.3 & 80.4 & 81.1 \\
        U-Transformer \cite{petit2021u} & 89.3 & 83.6 & 85.7 & 84.1 & 85.4 & 87.5 & 84.8 & 84.9\\
        DoDNet \cite{zhang2021dodnet} & 90.8 & 87.1 & 84.9 & 85.3 & 87.2 & 87.9 & 85.3 & 81.5 \\
        Clip-driven \cite{liu2023clip} & 91.1 & 87.6 & 85.1 & 86.4 & 88.5 & 89.1 & 82.4 & 83.9 \\
        UniSeg \cite{ye2023uniseg} & 92.3 & 86.5 & 87.7 & 86.3 & 88.3 & 93.1 & 84.9 & 85.1 \\
        UniverSeg \cite{butoi2023universeg} & 83.3 & 80.7 & 82.1 & 81.8 & 81.0 & 83.2 & 80.3 & 82.9\\
        \midrule
        Ours & 93.2 & \textbf{88.8} & \textbf{88.5} & 88.3 & \textbf{89.4} & \textbf{93.7} & 85.3  & \textbf{86.6} \\
        \bottomrule
    \end{tabularx}
\end{table}
\begin{figure}[t]
    \centering
    \includegraphics[width=1.0\linewidth]{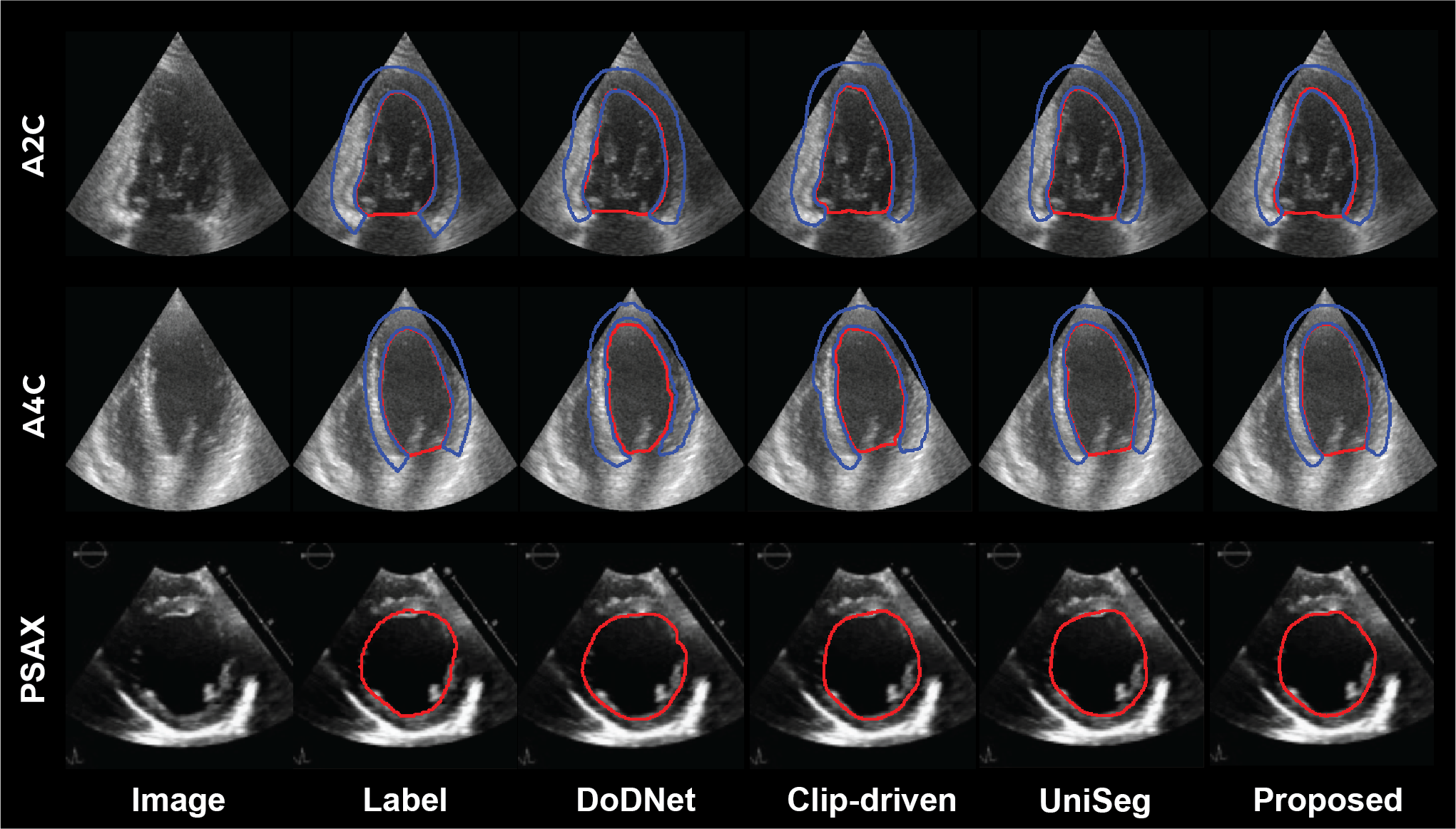}
    \caption{Qualitative visualization of segmentation results generated from our method and state-of-the-art  universal methods on representative image. \textcolor{red}{Red} and \textcolor{blue}{blue} represents LV$_{endo}$ and LV$_{epi}$, respectively.}
    \vspace{-5mm}
    \label{figure2}
\end{figure}

\section{Experiments and Results}
\begin{figure}[t]
    \centering
    \begin{minipage}[b]{0.35\textwidth}
        \centering
        \begin{tabular}{p{2.7cm}cc} % Adjust the width as per your requirement
            \hline
            \textbf{Encoder design} & \textbf{Dice [\%]} \\ \hline
            w/o text-encoder & 85.6               \\
            w/  One-hot & 88.6               \\
            \hspace{5mm}CLIP         & 88.8               \\
            \hspace{5mm}ClinicalBert & 89.6               \\ \hline
        \end{tabular}
        \captionof{table}{Comparison of mean model performance on different encoder design.}
        \label{table3}
    \end{minipage}%
    \hfill
    \begin{minipage}[b]{0.3\textwidth}
        \centering
        \begin{tabular}{p{1.3cm}c} % Adjust the width as per your requirement
            \hline
            \textbf{View} & \textbf{Dice [\%]} \\ \hline
            w/       & 89.4                                  \\
            w/o      & 89.6                                  \\ \hline
        \end{tabular}
        \captionof{table}{Comparison of model performance with and without explicit view information.}
        \label{table4}
    \end{minipage}%
    \hfill
    \begin{minipage}[b]{0.3\textwidth}
        \centering
        \includegraphics[width=\linewidth]{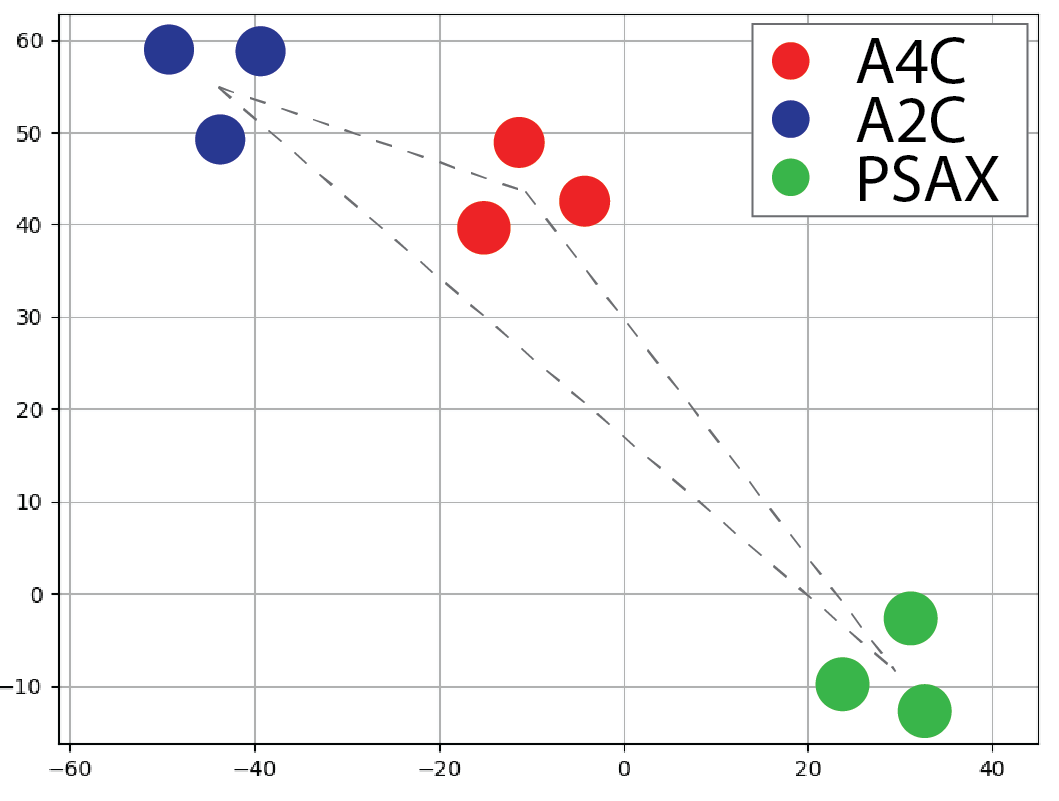}
        \caption{T-SNE visualization of prompt keys.}
        \label{figure3}
    \end{minipage}
\end{figure}

\textbf{Materials.} 
We evaluated the proposed method using three publicly available datasets \cite{leclerc2019deep}\cite{reddy2023video}\cite{ouyang2020video}. These datasets consist of 2D B-mode scan data annotated different cardiac chambers at end-diastole (ED) and end-systole (ES). The annotation includes the left ventricle endocardium (LV$_{\text{endo}}$) and the left ventricle epicardium (LV$_{\text{epi}}$) in apical two-chamber (A2C), apical four-chamber (A4C), and parasternal short-axis (PSAX) views. We followed a predefined split set as shown in Table \ref{tab1}.\\
\textbf{Implementation and Evaluation Metric}
To guarantee a fair comparison across all experiments, we standardized the training and testing settings. The experiments were conducted using PyTorch, with a consistent batch size of 5 over 100 epochs with Nvidia A100. We utilized the Unet architecture \cite{ronneberger2015u} as the backbone to incorporate our key components. For optimization, we employed the MADGRAD optimizer \cite{defazio2022adaptivity}, setting the learning rate to 1e-4. Images are resized to 224$\times$224 pixels with 16 frames and normalized to ensure a zero mean and unit variance. To enhance the robustness of our model, we apply various augmentation technique, including random flip, rotation within a range of $-$30 to +30, and shearing along x-y dimensions. We utilize the Dice Similarity Coefficient (DSC) for the evaluation of our model's performance. We compared our method across three scan views: A4C, A2C, and PSAX. These scan views were selected based on the currently available echo dataset. The performance of the models was evaluated on ED and ES cardiac phases, where annotations were available.\\
\textbf{Comparison study}
We present the performance of the proposed method for cardiac segmentation across different scan views. To the best of our knowledge, our method is the first approach capable of performing cardiac segmentation with view-agnostic input that can perform universal segmentation. We compared our proposed method under two different settings: 1) trained and tested on the same views noted as view-specific model approach, and 2) trained on all views and tested on all views noted as view-integrated model approach. We chose two methods, SwinUNETR \cite{hatamizadeh2021swin} and U-transformer \cite{petit2021u} for baseline segmentation model based on previous study \cite{kim2023medivista}. We also compare the performance of proposed method against universal models, including DoDNet \cite{reddy2023video}, CLIP-driven universal model \cite{liu2023clip}, UniSeg \cite{ye2023uniseg}, and UniverSeg \cite{butoi2023universeg} for cardiac segmentation tasks on three datasets. The CLIP-driven universal model \cite{liu2023clip} substitutes one-hot embeddings \cite{zhang2021dodnet} with CLIP text embeddings. UniSeg \cite{ye2023uniseg} employs a learnable prompt for generating task-specific embeddings. Additionally, our comparision includes a few-shot based universal segmentation model called UniverSeg \cite{butoi2023universeg}.\\
As depicted in Table \ref{table2}, our method exhibits the capability to generate excellent segmentation results even with view-agnostic input, as illustrated in Fig. \ref{figure2}. In comparison to baseline models with view-integrated approaches, our model demonstrates better performance compared to the view-specific based U-transfomrer except on LV$_{endo}$ (93.2 vs. 93.3) and LV$_{epi}$ (88.3 vs. 88.5) in A2C and A4C, respectively. Moreover, by incorporating prompts to adaptively integrate input view information, our model outperforms all universal models in delineating regions of interest (ROI) across all views. Since, our model adapts the prompt to match input view types, resulting in enhanced segmentation results and superior performance. Furthermore, our method improves mean segmentation performance compared to employing a few-shot segmentation method (89.64 vs 81.7). These experimental findings demonstrate that our approach effectively leverages adaptable prompts to produce superior segmentation results.\\
\textbf{Ablation study} To evaluate the effectiveness of the each component, we conducted an ablation study to quantify the impact of different elements in our model on segmentation performance. First, we evaluated performance metrics without text-encoder path, and then with text-encoder path employing various designs including one-hot encoding, as well as leveraging prior knowledge from language models including CLIP and ClinicalBert. In the absence of text-encoder path, which removes pixel-text alignment for auxiliary loss, we observed a degradation in performance (89.6 to 85.6) showing that an pixel-text alignment is crucial for our model. Additionally, we compared different types of text encoders trained on both natural and medical text. The findings indicate that the CLIP language model is not as effective in representing medical text compared to the ClinicalBert (88.8 vs. 89.6). Secondly, we assessed the classification performance based on selected prompt keys. We examined the assigned prompt keys with majority voting determined by $\arg\max_{G \in {A, B, C}} \sum_{i=1}^{3} \mathbb{I}(x_i \in G)$. T-SNE visualization \cite{van2008visualizing} is presented in Figure \ref{figure3}. We observed that the accuracy for distinguishing between apical views and parasternal views was 0.96. However, the accuracy for distinguishing between A2C and A4C was 0.54 and 0.6, respectively. Secondly, there is variability between view classes annotated by human readers due to ambiguous scan angles between A2C and A4C. In fact, the probe, guided by manual manipulation, often fails to accurately capture both the A2C and A4C angles when rotated from a single position. This is evident in Table \ref{table4}, where providing view information instead of selecting keys from the prompt pool resulted in a model performance of 89.4, lower than the performance of 89.6 achieved when view information was not provided.

\section{Conclusion}
In this study, we introduce an innovative prompt-driven universal echocardiography segmentation model that is capable of learning cardiac segmentation across different standard views using partially labeled data. This model incorporates the knowledge of pre-trained language models by aligning text representation with visual pixel data. We also suggest a prompt matching technique through a prompt pool to achieve view-agnostic echocardiography segmentation. Our research utilized three standard views to demonstrate the feasibility of our proposed model, highlighting its potential to be expanded to additional standard views to create a universal model for echocardiography. This method simplifies the process by eliminating the need for a separate step to identify views, thereby reducing the variability introduced by humans in selecting views for analysis in a patient's study. Extensive experiments on the echocardiography segmentation benchmark across various scan views have shown that our approach not only performs superiorly but also proves to be highly effective.

% ---- Bibliography ----
\bibliographystyle{splncs04}
\bibliography{ref}

\end{document}